\def\arxivversion{1}
\def\arxivversion{1}
\documentclass[pdflatex,sn-nature]{sn-jnl}

\input{sections/00_preamble.tex}

\ifdefined\arxivversion
\usepackage{bibunits}
\makeatletter
\let\arxivnatlinkstart\hyper@natlinkstart
\let\arxivnatlinkbreak\hyper@natlinkbreak
\let\arxivnatanchorstart\hyper@natanchorstart
\renewcommand{\hyper@natlinkstart}[1]{\arxivnatlinkstart{\@bibunitname.#1}}
\renewcommand{\hyper@natlinkbreak}[2]{\arxivnatlinkbreak{#1}{\@bibunitname.#2}}
\renewcommand{\hyper@natanchorstart}[1]{\arxivnatanchorstart{\@bibunitname.#1}}
\makeatother
\fi

\ifdefined\arxivversion\else
\usepackage{vruler}
\usepackage{etoolbox}
\patchcmd{\makevruler}{\tiny}{\normalfont\normalsize}{}{%
  \PackageError{aqua-paper}{Unable to configure line-number font}{Check the installed vruler package.}}
\setvruler[12bp][1][1][3][1][1.18\textwidth][26pt][-7pt][0.99\textheight]
\fi

\renewcommand{\figurename}{Fig.}
\hypersetup{bookmarksdepth=2,pdftitle={Self-Evolving Scientific Agent Designs Physically Reasoned White-Box Fluid Control}}
\setcitestyle{super,open={},close={}}
\floatstyle{plaintop}
\restylefloat{table}
\makeatletter
\renewcommand{\floatc@plain}[2]{{\fontsize{8}{9.5}\selectfont\raggedright
  \textbf{#1}\hspace{0.7em}#2\par}}
\makeatother

\title{Self-Evolving Scientific Agent Designs Physically Reasoned White-Box Fluid Control}

\author[1]{\fnm{Boai} \sur{Sun}}
\author[1]{\fnm{Wenjin} \sur{Guo}}
\author[1]{\fnm{Zongmin} \sur{Yu}}
\author*[1]{\fnm{Liu} \sur{Yang}}\email{yangliu@nus.edu.sg}
\affil[1]{\orgname{National University of Singapore}, \orgaddress{\country{Singapore}}}

\abstract{

While neural networks excel in autonomous control, their black-box nature makes control decisions difficult to interpret and diagnose in dynamic fluids. Here, we show how self-evolving scientific agents can design explicit, neural-network-free white-box controllers by iteratively interpreting simulation evidence, accumulating control knowledge and refining controller code. We demonstrate this approach on an underactuated two-joint swimmer navigating unsteady flows via joint angular accelerations. Starting from a target-blind propulsive controller, the agent gradually constructs key mechanisms, including travelling-wave propulsion, body-frame guidance, phase-selective steering, redirect bursts and adaptive relief. The resulting controllers reach targets and generalize across changes in target position, wake geometry, cylinder count, and inflow speed without revision. Moreover, 2D control priors transfer successfully to accelerate 3D adaptation. Our work demonstrates that self-evolving agents can autonomously design physically reasoned and generalizable white-box fluid control, showing a promising paradigm beyond traditional reinforcement learning and black-box neural network control.
}

\begin{document}
\maketitle

\ifdefined\arxivversion
\begin{bibunit}[sn-nature]
\fi

\phantomsection
\label{sec:introduction}

Traditionally, controlling complex, nonlinear physical systems has relied on hand-crafted feedback laws written directly as explicit control code. However, manually engineering such controllers for highly dynamic and chaotic environments, such as unsteady fluid flows, demands prohibitive expert human effort. It requires deep domain knowledge in fluid mechanics and control theory, mathematical simplification of intricate boundary conditions, and tedious, iterative trial-and-error to fine-tune high-dimensional heuristic parameters~\citep{slotine1991applied}. Because capturing nonlinear fluid-structure interactions analytically is exceptionally challenging, hand-crafting robust, high-performance control code often becomes an intractable engineering bottleneck.

To bypass this human-design cost, deep reinforcement learning (DRL) and neural-network-based policies emerged as the dominant paradigm, automating policy search through interactive data~\citep{rabault2019, verma2018, fan2020, vignon2023, novati2021, brunton2020machine, lagemann2026hydrogym}. Yet, DRL introduces its own formidable compromises: extreme sample inefficiency requiring expensive simulations, notorious training instability, and a heavy reliance on arbitrary reward shaping. Crucially, the resulting policies are black boxes~\citep{dulacarnold2021, banerjee2023}. Under physical perturbations, these neural networks are prone to catastrophic failures that are impossible for human engineers to inspect, diagnose, or systematically correct.

Recent breakthroughs in utilizing large language models (LLMs) as optimizers have driven remarkable progress in automated algorithm design and software synthesis~\citep{romeraparedes2024, yang2023, ma2024, novikov2025, boiko2023}.
Within control, LLMs have been used to generate robot
programs~\citep{liang2023}, tune controllers~\citep{guo2024controlagent},
translate PDE control specifications into optimization programs~\citep{soroco2025pde},
and synthesize interpretable policies through program search and numerical
optimization~\citep{bosio2025synthesizing, bosio2025combining}.
Weng~\citep{weng2026} introduces heuristic learning (HL), a paradigm in which
agents iteratively refine explicit policy code using environment feedback,
with MuJoCo locomotion serving as an illustrative application.
Here, we focus on white-box control in complex, unsteady fluid flows, where
propulsion and steering arise from nonlinear fluid-structure interactions.
Whether such a design loop can produce physically grounded controllers that
generalize across changing flow conditions remains an open question.
This process requires the agent to act much like a human scientist: deploying
candidate strategies within a physical simulation, actively diagnosing complex
dynamic failures (e.g., asymmetric propulsion or directional bias) from
multimodal evidence, and translating these observations into progressive
refinements.

To investigate this approach to white-box control design, we introduce a challenging non-linear computational fluid dynamics (CFD) control problem. A two-joint free swimmer is tasked with reaching varying spatial targets by modifying its body shape alone. This target-reaching problem is severely underactuated~\citep{fossen2011}. The controller has no direct access to global force, torque, or trajectory commands; it can only specify the angular accelerations of the body joints. Consequently, propulsion and steering must arise indirectly from the hydrodynamic response to the generated traveling body wave. Classical studies~\citep{lighthill1971, lighthill1970, triantafyllou2000, taylor2003, muller2002, borazjani2010, gazzola2014scaling} have shown that aquatic locomotor performance is highly sensitive to the timing of body waves, the distribution of curvature, and the resulting wake interactions. In this setting, converting a basic propulsive rhythm into reliable signed turning without target-specific behavioral rules becomes a stringent test of physical reasoning.

In this study, we use our Evolving Ensemble of Agents (EvE)
framework~\cite{yu2026eve} to develop white-box fluid controllers through the
co-evolution of design agents and executable control code.
Starting from a target-blind propulsive seed, the agents autonomously
constructed feedback controllers for navigating an unsteady
two-dimensional four-cylinder wake. We then froze the controllers and
challenged them with a generalization matrix varying target location, wake
geometry, obstacle number and flow speed. Without retraining, retuning or
case-specific branching, the controllers reached the target in the vast majority of tests.
With the policies and their design histories preserved as executable source code, both the
resulting controllers and the discovery process are directly inspectable. The
archived trace of the representative run reveals a structured architecture combining travelling-wave
propulsion, body-frame bearing guidance, phase-selective steering, redirect
bursts and load-conditioned relief. Finally, we transferred a retained
two-dimensional policy into a three-dimensional environment as the starting
point for continued evolution. Compared with evolution from the minimal seed,
the transferred experience stabilized the three-dimensional search and raised
the observed score ceiling, suggesting that the discovered control knowledge
can remain useful as physical complexity increases.

The results demonstrate how self-evolving agents can design white-box controllers
for the tested underactuated CFD problem. The retained source-code lineage and
simulation records connect successive design changes to observed behaviour,
while targeted analyses test the physical roles of the resulting control
structures. Together, these records make it possible to evaluate both controller
performance and the mechanisms through which control is achieved.

\section{Results}
\label{sec:results}

\subsection{Self-evolving agents for white-box fluid control}
\label{sec:results-framework}

Self-evolving agents construct white-box controllers through repeated source-code revision and
physical simulation. Agents receive a candidate controller, accumulated rollout
evidence and physical priors, then propose changes to the controller's logic
and parameters. A fixed CFD evaluator executes each candidate and returns a
score together with multimodal evidence. The resulting controller maps observed
state to joint commands through explicit code; the design-time agent is not
required during deployment. Figure~\ref{fig:scientific-agent-loop} summarizes
the co-evolution of design agents and controllers.

\begin{figure}[!hbp]
  \centering
  \includegraphics[width=\linewidth]{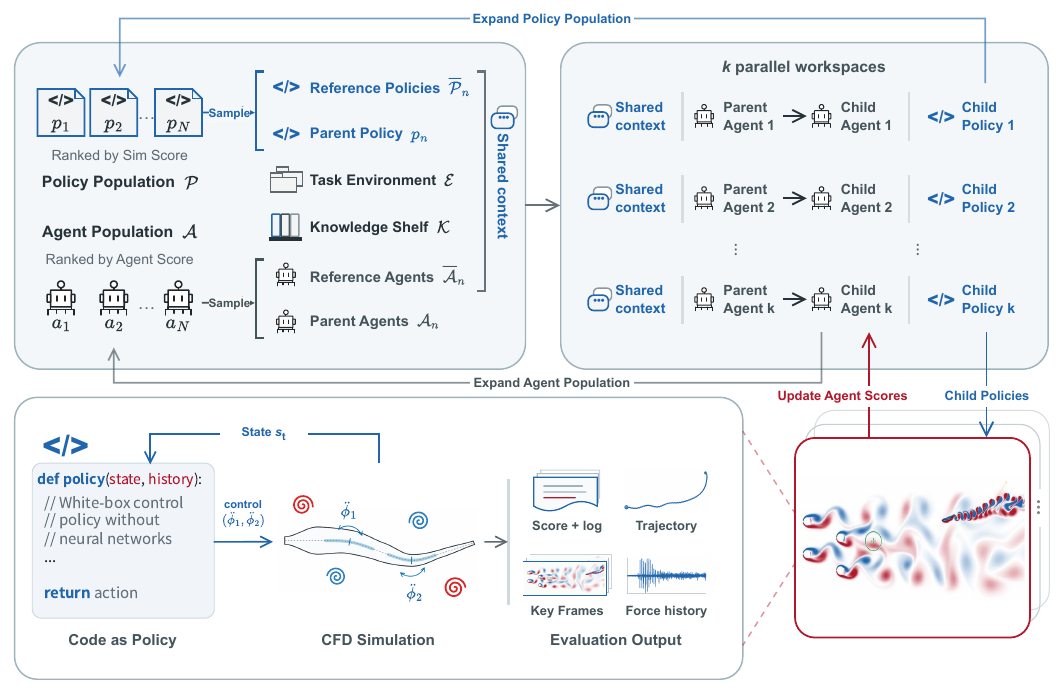}
\caption{\textbf{Self-evolving agents for white-box fluid control.}
  Design agents and executable control code evolve together.
  A parent policy
  and parent agents are sampled from their respective ranked populations.
  Reference policies, reference agents, the Knowledge Shelf, and
  the task environment form the shared context supplied to parallel agent workspaces.
  Each workspace returns a child agent and a child source policy. The child policy
  is evaluated through the fixed CFD interface, which returns a simulation
  score and a multimodal log. These evaluated policies, updated agent logs, and
  updated agent scores are then used to expand both populations for the
  next evolutionary iteration.}
  \label{fig:scientific-agent-loop}
\end{figure}

For the present task, each controller can command only the two joint angular
accelerations $(\ddot{\phi}_1,\ddot{\phi}_2)$. It cannot directly prescribe
the swimmer's position, heading, force or torque. Propulsion and steering must
therefore arise through body deformation and the resulting hydrodynamic
response. Agents interpret trajectories, body and target diagnostics, runtime
logs and selected flow-field states to formulate changes in control structure.
The Knowledge Shelf supplies established physical concepts that can inform
these changes; each proposed controller must still be tested in the CFD
environment.

The task environment and Knowledge Shelf remain fixed throughout an experiment.
Agents revise controller code, while simulation scores are assigned exclusively
by the fixed evaluator. Each evaluated controller is retained with its score
and rollout evidence, alongside the associated design records. These materials
connect proposed changes to executable code and subsequent physical outcomes,
allowing the development and operation of the controller to be examined.

Methods and Algorithm~\ref{alg:eve-control} describe the implementation.

\subsection{Autonomous design and generalization in two-dimensional flow}
\label{sec:results-2d}

To test whether self-evolving agents could design effective controllers in a highly unsteady
flow, we challenged the swimmer with
the two-dimensional wake-navigation benchmark shown in
Fig.~\ref{fig:structured-evolution}d.
Four staggered cylinders at the reference Reynolds number $Re=1000$ generated
an evolving vortex street.
Starting downstream and obliquely oriented
to the mean flow, the swimmer had to turn upstream, descend through the wake and
continually adjust to fluctuating fluid forces to reach a target in the wake
core. The task therefore required propulsion, steering and obstacle avoidance.

\begin{figure}[!t]
  \centering
  \includegraphics[
    width=0.98\linewidth
  ]{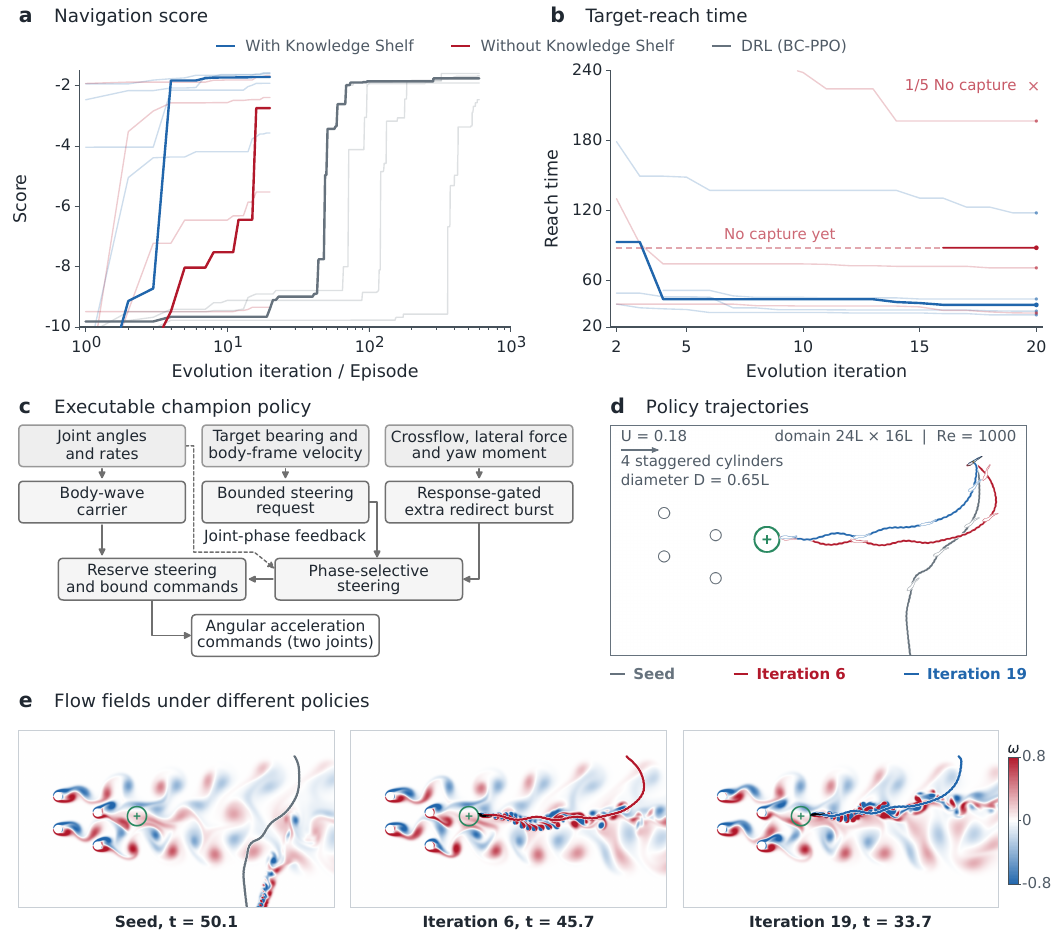}
  \caption{\textbf{Evolution of target-reaching policies in an unsteady wake.}
  \textbf{a}, Running-best score $s$ for self-evolving agents with and without the
  Knowledge Shelf over evolutionary iterations, and BC--PPO over training episodes.
  \textbf{b}, Best target reach time.
  Faint lines show five runs per condition; highlights trace the run with the
  median final value. Dashes indicate pre-capture status.
  \textbf{c}, Structure of the executable champion controller.
  \textbf{d}, Flow-field setup and trajectories for the Seed and evolved policies.
  \textbf{e}, Corresponding vorticity fields at the indicated times.}
  \label{fig:structured-evolution}
\end{figure}

The underactuated propulsive baseline (Seed) highlights the severity of this
problem: although it undulated forward, wake disturbances carried it off course
until it missed the target and left the domain
(Fig.~\ref{fig:structured-evolution}d,e). The agents established functional steering
early in the evolutionary run, but these initial successes had long and
variable capture times. They therefore marked acquisition of the task rather
than the end of optimization.

Across repeated searches, the retained navigation score improved
with the Knowledge Shelf, while successful-candidate capture times generally
decreased (Fig.~\ref{fig:structured-evolution}a,b).
To align the propulsive starting point for a fair
comparison, we initialized proximal policy optimization (PPO) by behavior
cloning the common seed controller (BC--PPO;
\hyperref[sec:methods-drl]{Methods}). Self-evolving agents with the Knowledge Shelf reached comparable
navigation scores within substantially fewer evolutionary iterations than the
number of training episodes required by BC--PPO
(Fig.~\ref{fig:structured-evolution}a).

The representative trajectories in Fig.~\ref{fig:structured-evolution}d,e
illustrate how evolutionary refinement reorganized the swimmer's route. Unlike the
drifting Seed and the winding intermediate policy, the final controller turned
directly into the wake core and reached the target along a much shorter path.

This more direct route did not compromise safety: both evolved
policies maintained more than $3.4L$ clearance from the cylinders.
The swimmer remained within its joint-angle limits despite frequent
saturation of joint rates and accelerations, showing that the policies actively
exploited the available dynamic envelope to overcome hydrodynamic forces.
Rather than uniformly reducing hydrodynamic loading, the
agent accepted transient forces as it reshaped the route.
\hyperref[sec:results-control-strategy]{The next subsection} examines how the
controller evolved in the representative run.
Figure~\ref{fig:structured-evolution}c shows the code structure of the
Iteration-19 champion controller. Its 244-line Julia implementation maps sensory
inputs to bounded angular-acceleration commands for the two joints.

To confirm that the evolved controllers provide true closed-loop feedback
rather than a memorized trajectory, we evaluated the fixed
policies in the setting used during evolution and in 12 generalization
tests conducted afterwards. No retraining, tuning or case-specific branching
was introduced for these evaluations.
The one-factor-at-a-time matrix
(Table~\ref{tab:generalization-matrix}) independently perturbed target
coordinates, rear-row geometry, cylinder count and inflow speed, thereby
changing both obstacle--wake structure and characteristic convective
timescales while leaving the other three variables fixed.

The controller shown in the code flowchart
(Fig.~\ref{fig:structured-evolution}c) reached the target in the original
setting and in every generalization test, across changes in target position,
rear-row geometry, cylinder count and inflow speed. The other retained
policies also reached the target in most generalization tests.
Because the policy source and parameters were
fixed, the distinct successful responses demonstrate closed-loop adaptability
rather than case-specific specialization.
By contrast, evaluations of frozen DRL (BC--PPO)
policies showed poor target-capture performance
under the same perturbations
(Table~\ref{tab:generalization-matrix}).
Representative responses from cases
a1, b3, c1 and d3 are shown in Fig.~\ref{fig:generalization}.

\noindent\begin{minipage}{\linewidth}
\setlength{\intextsep}{3pt}
\begin{figure}[H]
  \centering
  \includegraphics[
    width=0.98\linewidth,
    keepaspectratio
  ]{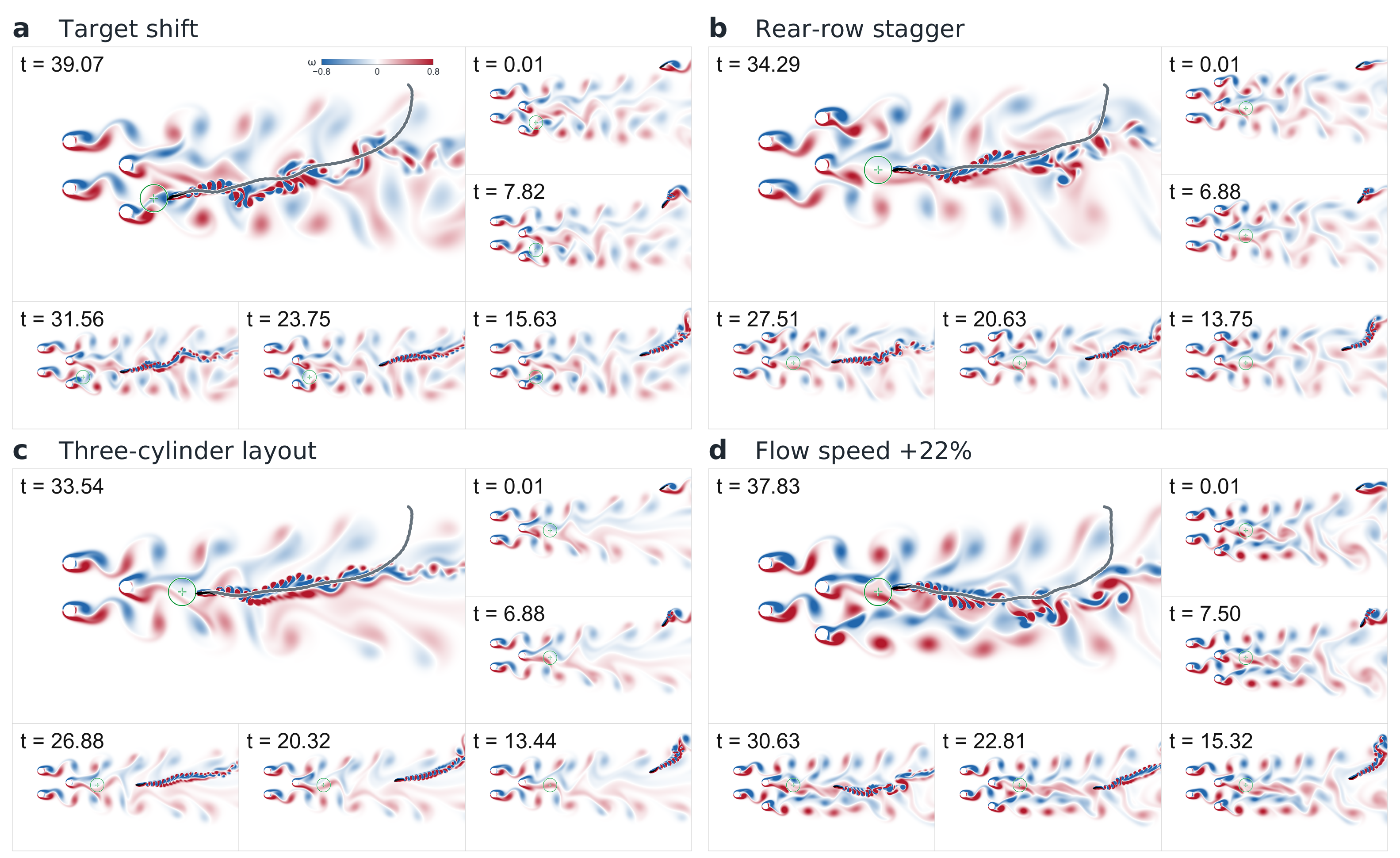}
  \caption{\textbf{Representative two-dimensional generalization tests with a
  fixed policy.} Vorticity fields from the
  generalization matrix for \textbf{a}, a target shifted to
  $(x_t,y_t)/L=(7.5,8)$ (a1); \textbf{b}, a rear row with its vertical stagger
  reversed relative to the front row, $\mathcal R/L=(6;9.75,12.25)$ (b3);
  \textbf{c}, a three-cylinder layout retaining $C_1$, $C_2$ and $C_4$ (c1);
  and \textbf{d}, an increased inflow speed $U=0.22$ ($+22.2\%$; d3).
  The same retained policy was used without retraining, tuning or case-specific
  branching.}
  \label{fig:generalization}
\end{figure}

\begin{table}[H]
  \centering
  \caption{\textbf{Fixed-policy generalization matrix.}
  Each subtable applies one-factor-at-a-time environmental perturbations while
  the policies and the other three variables remain fixed.}
  \label{tab:generalization-matrix}
  \begingroup
  \fontsize{7}{9}\selectfont
  \setlength{\tabcolsep}{0.6pt}
  \renewcommand{\arraystretch}{1.02}
  \setlength{\aboverulesep}{0.8pt}
  \setlength{\belowrulesep}{1.0pt}

  \begin{minipage}[t]{0.49\linewidth}
    \centering
    \textbf{a. Target coordinate}\par\vspace{2pt}
    \begin{tabular*}{\linewidth}{
      @{\extracolsep{\fill}}cccccc@{}
    }
      \toprule
      \multirow{2}{*}[-1pt]{Test} & \multirow{2}{*}[-1pt]{$(x_t,y_t)/L$}
      & \multicolumn{3}{c}{Agent} & DRL \\
      \cmidrule(lr){3-5}\cmidrule(l){6-6}
      & & Reach & $t_{\mathrm{champ}}$ & $s_{\mathrm{champ}}$ & Reach \\
      \midrule
      Original & $(9,9.5)$ & 5/5 & 33.73 & $-1.5941$ & 5/5 \\
      a1 & $(7.5,8)$ & 5/5 & 39.07 & $-1.8232$ & 0/5 \\
      a2 & $(7.5,11)$ & 5/5 & 36.98 & $-1.7216$ & 0/5 \\
      a3 & $(10.5,9.5)$ & 5/5 & 31.15 & $-1.4407$ & 0/5 \\
      \bottomrule
    \end{tabular*}
  \end{minipage}\hfill
  \begin{minipage}[t]{0.49\linewidth}
    \centering
    \textbf{b. Rear-row configuration}\par\vspace{2pt}
    \begin{tabular*}{\linewidth}{
      @{\extracolsep{\fill}}cccccc@{}
    }
      \toprule
      \multirow{2}{*}[-1pt]{Test} & \multirow{2}{*}[-1pt]{$\mathcal R/L$}
      & \multicolumn{3}{c}{Agent} & DRL \\
      \cmidrule(lr){3-5}\cmidrule(l){6-6}
      & & Reach & $t_{\mathrm{champ}}$ & $s_{\mathrm{champ}}$ & Reach \\
      \midrule
      Original & $(6;7.25,9.75)$ & 5/5 & 33.73 & $-1.5941$ & 5/5 \\
      b1 & $(4.5;7.25,9.75)$ & 4/5 & 34.35 & $-1.5872$ & 0/5 \\
      b2 & $(7.5;7.25,9.75)$ & 5/5 & 32.87 & $-1.5834$ & 0/5 \\
      b3 & $(6;9.75,12.25)$ & 5/5 & 34.29 & $-1.5939$ & 0/5 \\
      \bottomrule
    \end{tabular*}
  \end{minipage}\par\vspace{4pt}

  \begin{minipage}[t]{0.49\linewidth}
    \centering
    \textbf{c. Number and layout of cylinders}\par\vspace{2pt}
    \begin{tabular*}{\linewidth}{
      @{\extracolsep{\fill}}cccccc@{}
    }
      \toprule
      \multirow{2}{*}[-1pt]{Test} & \multirow{2}{*}[-1pt]{\shortstack{$N_c$\\(retained $C_i$)}}
      & \multicolumn{3}{c}{Agent} & DRL \\
      \cmidrule(lr){3-5}\cmidrule(l){6-6}
      & & Reach & $t_{\mathrm{champ}}$ & $s_{\mathrm{champ}}$ & Reach \\
      \midrule
      Original & $4;\{1,2,3,4\}$ & 5/5 & 33.73 & $-1.5941$ & 5/5 \\
      c1 & $3;\{1,2,4\}$ & 4/5 & 33.54 & $-1.5988$ & 0/5 \\
      c2 & $2;\{2,4\}$ & 3/5 & 35.24 & $-1.6253$ & 0/5 \\
      c3 & $1;\{4\}$ & 4/5 & 35.22 & $-1.6149$ & 0/5 \\
      \bottomrule
    \end{tabular*}
  \end{minipage}\hfill
  \begin{minipage}[t]{0.49\linewidth}
    \centering
    \textbf{d. Flow speed}\par\vspace{2pt}
    \begin{tabular*}{\linewidth}{
      @{\extracolsep{\fill}}cccccc@{}
    }
      \toprule
      \multirow{2}{*}[-1pt]{Test} & \multirow{2}{*}[-1pt]{$U$}
      & \multicolumn{3}{c}{Agent} & DRL \\
      \cmidrule(lr){3-5}\cmidrule(l){6-6}
      & & Reach & $t_{\mathrm{champ}}$ & $s_{\mathrm{champ}}$ & Reach \\
      \midrule
      Original & 0.18 & 5/5 & 33.73 & $-1.5941$ & 5/5 \\
      d1 & 0.16 & 5/5 & 32.80 & $-1.5704$ & 0/5 \\
      d2 & 0.20 & 3/5 & 34.92 & $-1.6250$ & 0/5 \\
      d3 & 0.22 & 4/5 & 37.83 & $-1.7350$ & 0/5 \\
      \bottomrule
    \end{tabular*}
  \end{minipage}\par\vspace{4pt}
  \begin{minipage}{0.98\linewidth}
    \fontsize{7.5}{9}\selectfont
    \textit{Note:} Agent denotes white-box controllers generated by self-evolving agents; DRL denotes BC--PPO policies. $\mathcal R/L=(x_r;y_{r1},y_{r2})$ denotes the
    rear-row configuration. Original denotes the setting used during
    evolution: $(x_t,y_t)/L=(9,9.5)$,
    $\mathcal R/L=(6;7.25,9.75)$, $N_c=4$, $D/L=0.65$ and $U=0.18$.
    Its cylinder centres are
    $C_1=(3,8.5)L$, $C_2=(3,11)L$, $C_3=(6,7.25)L$ and
    $C_4=(6,9.75)L$; braces list the retained cylinders.
    Each test uses a fingerprint-matched prewarm state.
    $t_{\mathrm{champ}}$ and $s_{\mathrm{champ}}$ report generalization results
    for the fixed champion policy.
    Reach ($x/5$) counts target captures among five frozen policies per method,
    each verified in the original setting.
  \end{minipage}
  \endgroup
\end{table}

\end{minipage}

\subsection{Evolution of the white-box control strategy}
\label{sec:results-control-strategy}

In evolving the controller shown in
Fig.~\ref{fig:structured-evolution}c, the agent retained the Seed's travelling body wave and added
feedback that improved steering and the final approach to the target.
The minimal seed used a naive oscillator and a phase lag
between the joints to produce
forward motion, but it was target-blind and therefore had no route information
with which to steer through the wake. Later policies used the bearing more
directly to bend the body towards the target and combined it with measured
body-frame motion to allocate steering authority.

As evolution continued, the controller also began to use recent observations
of the target and measurements of the surrounding flow. Comparing the current
bearing with its recent values helped the controller distinguish a steering
error that lasted over time from a short change caused by the wake.
Target-aligned crossflow corroborated by lateral force, an assisting
hydrodynamic moment, or high joint speed could each withdraw only surplus
redirect authority while leaving the base steering intact. The controller also
redistributed steering across the body-wave phase and supplied a temporary
burst when a large redirection remained unmet. Each change was limited and
added to the existing wave rather than replacing it with a separate action
sequence.

This staged accumulation of control structure is quantified
in Fig.~\ref{fig:physical-mechanisms}a. By Iteration~19, the non-comment source-code line count,
parameter count and number of feedback signals had increased to 7.3,
4.2 and 3.5 times their values in the minimal seed, respectively. These
increases did not add actuation degrees of freedom: every policy still returned
only the same two joint angular accelerations. The growth therefore reflects an
increasingly structured use of state and flow evidence rather than more direct
control authority.

Because each retained policy remains an executable source program, this
lineage can be followed in the archived source policies and simulation
records. Unlike BC--PPO's neural-network policy, the evolved controller exposes
its feedback logic directly in executable source code.
The Knowledge Shelf supplied established ideas about fish propulsion,
phase lag, steering and responses to fluid loads, but the controller did not
copy a complete design. Instead, the records show how these ideas were
connected to quantities available in the simulation, such as target bearing,
body-frame velocity, crossflow, lateral force and hydrodynamic moment, and then
expressed as changes in the angular accelerations of the two joints. The same
records link a proposed change to the edited source code and to the result of
the next simulation, so readers can inspect how the controller developed.
Selected policies and their implemented feedback pathways are summarized in
Supplementary Table~2.
The contrast in Fig.~\ref{fig:structured-evolution}a,b shows the
practical value of this context: the Knowledge Shelf improved
average search efficiency and learning stability, yielding successful controllers
in every run, whereas searches without it were less reliable. Together, this comparison and the archived
lineage show how reusable physical priors supported efficient policy development
without specifying a complete controller: each idea still had to be expressed
in executable feedback and retained through CFD evaluation.

\subsection{Physical mechanisms underlying adaptive wake navigation}
\label{sec:results-physical-mechanisms}

Having established the code-level lineage, we next tested how
the controller shown in Fig.~\ref{fig:structured-evolution}c
produced adaptive motion in the wake. We replayed
policies in this lineage from matched instantaneous wake states and paired
this controller with one-family-at-a-time removals, without retraining or retuning. The
analysis proceeds from realized motion and surface loading
(Fig.~\ref{fig:physical-mechanisms}b,c), through same-state command
counterfactuals and yaw moment (Fig.~\ref{fig:physical-mechanisms}d), to
wake-state robustness and full-rollout ablations
(Fig.~\ref{fig:physical-mechanisms}e,f).

\begin{figure}[!tbp]
  \centering
  \includegraphics[width=\linewidth]{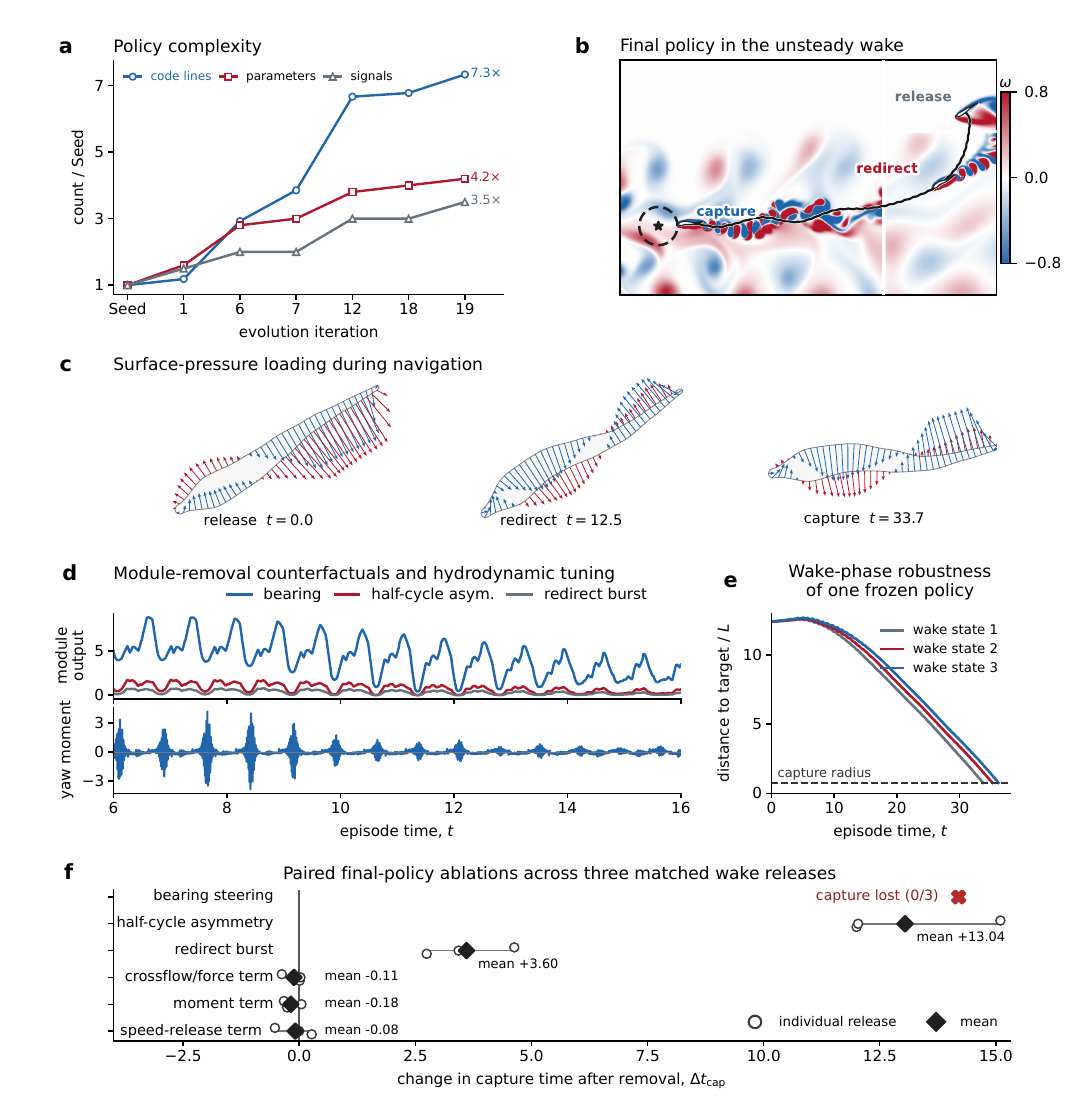}
  \caption{\textbf{Physical mechanisms of adaptive wake navigation.}
  \textbf{a}, Growth of the archived policy architecture from the target-blind
  Seed in a single evolutionary run. \textbf{b,c}, Final-policy trajectory and signed surface-pressure
  loading at release, redirection and capture.
  \textbf{d}, Nested same-state command counterfactuals and the
  full-policy hydrodynamic yaw moment during redirection.
  \textbf{e}, Target-distance histories of the final policy from three
  instantaneous wake states. \textbf{f}, Capture-time changes under
  final-policy ablations across three matched wake releases.}
  \label{fig:physical-mechanisms}
\end{figure}

The trajectory and loading views in
Fig.~\ref{fig:physical-mechanisms}b,c follow the swimmer from release through a
large redirection and into capture while the travelling deformation persisted. At
the same instants, the signed
surface-pressure vectors changed with both body posture and the encountered
wake, placing time-dependent loads at different moment arms along the body.
The target-guided command was therefore realized through deformation and
fluid loading, rather than as a prescribed global force or trajectory. These
snapshots establish the physical pathway, while the time-resolved
counterfactuals are needed to identify which control layers created the turn.

The rapid redirection is the central mechanism (Fig.~\ref{fig:physical-mechanisms}d).
The upper traces compare the full clipped joint command with same-state
counterfactual commands and are nested rather than additive: removing bearing
suppresses the entire target-guided pathway, removing half-cycle asymmetry
retains mean bearing steering but eliminates both the base and burst phase
redistribution, and removing the redirect burst retains the base asymmetry.
This hierarchy converts bearing error into a signed steering request,
concentrates that request in the half-cycle already favourable to the turn,
and supplies a transient surplus when redirection remains unmet. The resulting
phase-selective deformation produces the pressure loading and oscillatory yaw
moment shown below. The rollout lineage gives the same chain a performance
scale: Iteration~6 reorganized bearing steering through course response and
authority allocation and captured from all three releases; the isolated
half-cycle addition at Iteration~7 made capture roughly one-fifth faster
($19.6\%$); and the response-scheduled burst present by Iteration~12 provided
a smaller further acceleration. Thus the burst sharpens an already reliable
turn rather than replacing the bearing-to-phase steering mechanism.

The same controller also reached the target when released into three
distinct instantaneous wake states, without phase-specific retuning
(Fig.~\ref{fig:physical-mechanisms}e). The distance histories remain close but
not identical, showing that the policy did not replay one open-loop path:
wake-induced differences in body motion and loading were continually folded back
into the target-guided, phase-selective command. Wake-phase tolerance is
therefore a property of the complete closed loop.

\enlargethispage{\baselineskip}
Full-rollout ablations in
Fig.~\ref{fig:physical-mechanisms}f resolved the causal roles of the evolved
controller layers. Removing bearing steering eliminated all three captures;
removing half-cycle asymmetry preserved capture but caused the largest delay
(about 13 time units); and removing the redirect burst caused a smaller delay
of 3.6 time units. The later terms form a coherent
response-relief layer: corroborated target-aligned crossflow and lateral force,
an already assisting yaw moment, or rapid joint motion each withdraw surplus
burst authority. Consistent with this response-relief role,
their isolated removals preserved capture and changed reach time only
marginally, although the terms recur in winning policies along this lineage. Much
as a gene's phenotypic effect depends on genomic context, the contribution of
these terms may emerge through interactions with the surrounding controller
rather than as a separable gain in a single endpoint. A stabilizing regulator
can temper aggressive actuation and incur a small penalty under a target-reach
objective that rewards rapid arrival, yet help the controller better generalize to
complex, unsteady flows. Together, these results show how the
evolved controller uses target-guided commands to produce phase-selective
deformation, pressure loading and yaw moment, leading to wake-robust capture.
This explicit physical chain makes the controller physically
interpretable and distinguishes its source-code representation from a
neural-network policy whose state--action mapping is encoded in learned
weights.

\subsection{Transferring two-dimensional control priors to three dimensions}
\label{sec:results-3d}

After establishing a generalizable controller and identifying
its physical mechanisms in two dimensions, we used three-dimensional CFD to
test whether a retained two-dimensional controller could support further
design in three dimensions. The three-dimensional task retained the \(x\)--\(y\) navigation layout
in still water while adding a finite \(1.5L\) span, spanwise flow structures
and a deforming three-dimensional fluid--body interface. To make repeated
evaluations tractable, a \(4L\times3L\times1.5L\) computational window followed
the swimmer within the \(24L\times16L\) virtual world, as detailed in Methods.
Figure~\ref{fig:three-dimensional-adaptation}a shows the complete
release-to-capture trajectory and the window position at \(t=12\), showing that
the moving-window formulation supported a route extending beyond the window's
streamwise length. We compared three independent experiments initialized from
the retained two-dimensional policy with three initialized from the minimal
seed, each followed by 40 iterations of continued agent evolution in three dimensions.

A retained three-dimensional champion evolved from the
two-dimensional policy redirected the swimmer and reached the target while
retaining an undulatory gait
(Fig.~\ref{fig:three-dimensional-adaptation}b). Snapshots at
\(t=3\), \(t=9\), \(t=12\) and target reach at \(t=18.65\) trace the same
rollout. Immediately after release at \(t=3\), vortical structures began to
form behind the undulating body. By \(t=9\), their visibly wider streamwise
spacing accompanied the swimmer's acceleration, while the main \(t=12\) frame
shows the developed three-dimensional vortical field around the body and along
its wake. At target reach, the shorter wake spacing accompanied deceleration on
the final approach. Together, the frames show the swimmer completing the full
route to target reach while the computational window followed its motion.

Figure~\ref{fig:three-dimensional-adaptation}c directly
compares continued three-dimensional evolution initialized from the retained
two-dimensional policy with evolution from the minimal seed. Using the
untransformed distance--time integral defined in
Eq.~\eqref{eq:navigation-score}, all three experiments initialized from the
two-dimensional policy entered a narrow high-score band early and remained
closely grouped. Evolution from the minimal
seed was slower on average and varied substantially across experiments, with
outcomes ranging from failure to produce a target-reaching policy within the
available budget to a best score that remained below the ceiling attained after
transfer.

The three-dimensional comparisons therefore show that
initialization from the retained two-dimensional policy improves search
reliability and raises the observed score ceiling. Despite the added fluid--structure
complexity of the three-dimensional setting, control experience acquired in
two dimensions transferred effectively and provided a stable basis for
continued evolution in more complex environments.

\begin{figure}[!tbp]
  \centering
  \includegraphics[width=\linewidth]{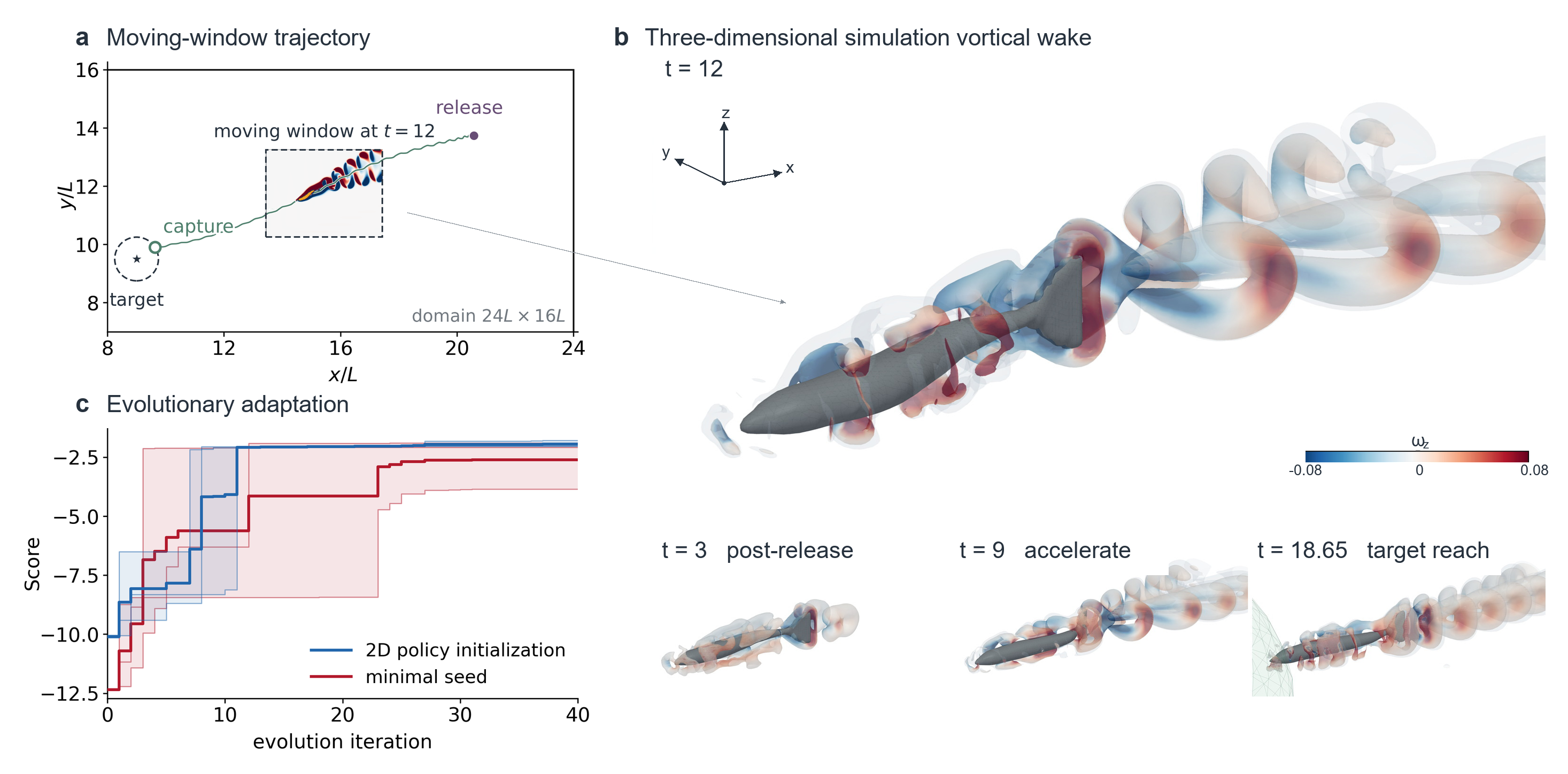}
  \caption{\textbf{Transferring 2D experience improves 3D adaptation.}
  \textbf{a}, Representative target-reaching
  trajectory in the moving-window domain, with the local mid-plane vortical
  wake shown at $t=12$. \textbf{b}, Three-dimensional swimmer and vortical wake
  at $t=12$, with additional frames spanning early motion to target reach.
  \textbf{c}, Best-score evolution for three independent experiments per
  condition; thick lines denote the mean, thin lines individual
  experiments and shading the observed range.}
  \label{fig:three-dimensional-adaptation}
\end{figure}

\section[Discussion]{Discussion}
\label{sec:discussion}

Our work demonstrates that self-evolving scientific agents can autonomously
design physically reasoned, inspectable white-box controllers for the fluid
navigation problem studied here, offering a promising paradigm beyond
traditional reinforcement learning and black-box neural-network
control~\citep{rabault2019, verma2018, banerjee2023}. Beyond refining controller
code, the agents revise and inherit reusable control-design knowledge, allowing
experience from earlier simulations to shape subsequent design decisions.
This approach separates design-time intelligence from run-time execution:
reasoning and code generation occur during off-line evolution, while the
resulting controller runs as a standalone program without language-model or
neural-network inference. This separation may facilitate deployment on
resource-constrained platforms, including autonomous underwater vehicles and
bio-inspired soft robots~\citep{fossen2011, katzschmann2018}.

The generalization results suggest that explicit feedback structures based on
relative target geometry, body state and hydrodynamic loads can remain
effective across changing flow conditions. The frozen controllers reached
targets in most tests involving changes in target position, wake geometry,
cylinder count and inflow speed, with the retained champion succeeding across
the full test matrix without retuning or revision. Frozen DRL policies showed
poor target-capture performance under the same perturbations. The explicit
controller code makes it possible to investigate how its feedback structures
respond to these changes. Matched-state physical analyses and rollout ablations
connect target-bearing guidance and phase-selective steering to deformation,
pressure loading and turning, while identifying smaller isolated effects from
other feedback terms. These analyses make the physical roles of individual
control structures testable; establishing which structures account for
robustness across different environments remains a question for further study.

The modularity of explicit source code also supports reuse as physical
complexity increases. Moving from two-dimensional to three-dimensional flows
introduces spanwise flow structures and a deforming three-dimensional
fluid--body interface. A retained two-dimensional controller provides a
structured starting point for continued evolution in this setting. Across
three independent experiments per condition, initialization from this
controller reached a high-score regime earlier, produced more consistent
outcomes and attained a higher observed score ceiling than initialization from
the minimal seed. These results support transferring executable control
structures, such as body-frame target guidance and travelling-wave
coordination, as priors for further design.

Several directions remain open for future exploration. First, controller
evaluation in the present study relies entirely on high-fidelity CFD
simulations. Whether agents can learn to autonomously use or train surrogate
models to reduce simulation cost is an important question for further work.
Second, the present approach provides physical interpretations of how the
generated controllers operate. An interesting next step is whether agents can
formalize this understanding mathematically, translating observed mechanisms
into models with explicit assumptions and deriving conditions under which
they remain effective. Such formalization could support analytical predictions
of controller behaviour and, ultimately, theoretical guarantees. Finally, the present results are
limited to simulation. How the controllers perform in real-world fluid
systems, and whether the acquired control strategies and design knowledge
can support simulation-to-real transfer, remain to be established.

\section{Methods}
\label{sec:methods}

\subsection{Implementation of self-evolving agents for fluid control}
\label{sec:methods-adc}

Our Evolving Ensemble of Agents (EvE) framework~\cite{yu2026eve}
provides the population-based search mechanism;
its use for controller design combines an editable controller-code interface,
a fixed CFD evaluator, multimodal rollout evidence and the Knowledge Shelf.
The deployed output is a standalone white-box controller.

The search state comprises a policy population $\mathcal{P}$ and an agent
population $\mathcal{A}$. Each policy tuple $(\pi,l^{\pi},v^{\pi})$ contains
executable controller code, a multimodal evaluation log and a simulation score.
Each agent tuple $(a,l^a,v^a)$ comprises an agent $a$ conditioned on its
inherited control-design knowledge, an associated log $l^a$ and an Elo score
$v^a$. This knowledge contains reusable lessons, failure modes and conditions
for their application, distilled from simulation evidence. We use
\emph{self-evolving agents} to describe agents whose inherited design knowledge
is revised and selected across iterations, thereby shaping subsequent
controller design. The underlying language model and execution framework
remain fixed. The initial populations $\mathcal{P}_0$ and $\mathcal{A}_0$
provide the Iteration-0 bootstrap. Algorithm~\ref{alg:eve-control} specifies the
population-level procedure; the operations below describe its use for fluid control.

\begin{algorithm}[H]
\small
\caption{Co-evolution of control-design agents and controller code}
\label{alg:eve-control}
\DontPrintSemicolon
\SetAlgoLined
\SetKwComment{Comment}{$\triangleright$\ }{}
\KwIn{Task environment $\mathcal{E}$, Knowledge Shelf
$\mathcal{K}$; task evaluator $f$; initial policy
population $\mathcal{P}_0$; initial agent population $\mathcal{A}_0$; total
iterations $T$.}
\KwOut{Updated populations $\mathcal{P}$ and $\mathcal{A}$.}

$\mathcal{P} \gets \mathcal{P}_0$; $\mathcal{A} \gets \mathcal{A}_0$\;

\For{$n \gets 1$ \KwTo $T$}{
  $\mathcal{A}_n,\overline{\mathcal{A}}_n
    \gets \operatorname{Sample}(\mathcal{A})$
  \Comment*[r]{sample parent and reference agents}
  $\mathcal{A}_n=\{(a_i,l_i^a,v_i^a)\}_{i\in I_n}$;
  $\overline{\mathcal{A}}_n=
    \{(a_r,l_r^a,v_r^a)\}_{r\in R_n}$\;
  $p_n,\overline{\mathcal{P}}_n
    \gets \operatorname{Sample}(\mathcal{P})$
  \Comment*[r]{sample parent and reference policies}
  $p_n=(\pi_n,l_n^{\pi},v_n^{\pi})$;
  $\overline{\mathcal{P}}_n=
    \{(\pi_j,l_j^{\pi},v_j^{\pi})\}_{j\in J_n}$\;

  \For{$i\in I_n$ \textnormal{ in parallel}}{
    $(\widehat{\pi}_i,\widehat{a}_i,\widehat{l}_i^a)
      \gets a_i(p_n,\overline{\mathcal{P}}_n,
                  \overline{\mathcal{A}}_n,\mathcal{E},\mathcal{K})
                  $
    \Comment*[r]{generate child policy and agent}
    $(\widehat{l}_i^{\pi},\widehat{v}_i^{\pi})
      \gets f(\widehat{\pi}_i)$
    \Comment*[r]{return multimodal log and score}
    $l_i^a \gets l_i^a \cup \widehat{l}_i^a$
    \Comment*[r]{update agent logs}
  }

  $\mathcal{P} \gets \mathcal{P} \cup
    \{(\widehat{\pi}_i,\widehat{l}_i^{\pi},
        \widehat{v}_i^{\pi})\}_{i\in I_n}$
  \Comment*[r]{expand policy population}
  $\{v_i^a\}_{i\in I_n} \gets
    \operatorname{EloUpdate}
    (\{\widehat{v}_i^{\pi}\}_{i\in I_n},\{v_i^a\}_{i\in I_n})$
  \Comment*[r]{update agent scores}
  $\mathcal{A} \gets \mathcal{A} \cup
    \{(\widehat{a}_i,\widehat{l}_i^a,
        \widehat{v}_i^a=v_i^a)\}_{i\in I_n}$
  \Comment*[r]{expand agent population}
}

\Return{$(\mathcal{P},\mathcal{A})$}\;
\end{algorithm}

\paragraph{Sample parent and reference populations.}
At iteration $n$, EvE samples parent and reference agents from
$\mathcal{A}$ and a parent policy $p_n$ together with reference policies from
$\mathcal{P}$. The parent policy is the source controller to be modified. The
reference policies and agents provide population-level context rather than
additional executable branches in the controller. The same sampled parent
policy and shared context are supplied to every parallel workspace, while each
workspace is assigned its sampled parent agent.

\paragraph{Generate a child agent--policy pair.}
Within workspace $i$, agent $a_i$ receives the parent policy, reference
policies, reference agents, task environment and Knowledge Shelf. From these
inputs it produces a child source policy $\widehat{\pi}_i$, a child agent
$\widehat{a}_i$ and a child agent-log entry $\widehat{l}_i^a$. Physical diagnosis
and source-code revision occur inside this generation operation and are
recorded as reasoning in the agent log, not as separate population nodes. The
Knowledge Shelf provides reference context; consultation requirements are
specified in
Supplementary Section~B. It does not directly edit or score a policy.

\paragraph{Evaluate the child policy.}
The fixed task evaluator executes $\widehat{\pi}_i$ in the CFD--FSI
environment. The policy can command only the two joint angular accelerations
$(\ddot{\phi}_1,\ddot{\phi}_2)$; translation and turning remain consequences
of body deformation and hydrodynamic reaction. The evaluator returns exactly
two policy-level artifacts,
$(\widehat{l}_i^{\pi},\widehat{v}_i^{\pi})$: a multimodal log and a simulation
score. The multimodal log is the native record of the rollout, including
runtime policy logs, termination and trajectory summaries, body and target
diagnostics, and selected flow-field states. It is returned directly with the
score as the evaluation output and remains attached to the evaluated policy.

\paragraph{Update agent logs and scores.}
The generated agent-log entry is appended to the parent agent log,
$l_i^a\leftarrow l_i^a\cup\widehat{l}_i^a$. Once all parallel policies have
been evaluated, their externally assigned simulation scores
$\widehat{v}_i^{\pi}$ are used by the Elo update to revise agent scores. This
separates the agent's textual reasoning record from the physical evaluator's
performance judgement.

\paragraph{Expand the policy and agent populations.}
Each evaluated policy is inserted into $\mathcal{P}$ together with its own
multimodal log and score. Each generated agent is inserted into $\mathcal{A}$
together with its updated agent log and revised agent score. The two expanded
populations supply the parent and reference samples for the next iteration.
This implementation retains explicit white-box controllers and the paired
artifacts needed to inspect their development: multimodal simulation logs,
agent logs and scores. Controller evolution proceeds through revisions to
executable source code without training or deploying a
neural-network policy.

The Knowledge Shelf in the shared context contains established priors from
fish locomotion~\citep{lighthill1960, lighthill1971, taylor1952,
triantafyllou2000, sfakiotakis1999, anderson1998, taylor2003, floryan2018,
domenici1997} and robotic feedback control~\citep{ijspeert2008, fossen2011,
katzschmann2018}. These priors describe general mechanisms rather than a
controller for the present body. Any borrowed concept must still be translated
into the available state variables and joint-acceleration interface, written as
readable source code and accepted by a subsequent CFD evaluation.

Finally, the edit surface is intentionally constrained. Every generated policy
must return finite $(\ddot{\phi}_1,\ddot{\phi}_2)$ through a unified controller
and must avoid shortcuts such as hidden time tracking, random numbers or
case-name branching. A policy is retained with the score and multimodal log
produced by the same physics-based evaluation protocol used for every other
candidate.

The launch prompt and the workspace instructions provided to each agent,
together with the model and execution settings, the worker
context, population sampling, editable surface and retained evidence, are
reported in Supplementary Section~B.

\subsection{Fluid control simulation}
\label{sec:methods-dogfish-simulation}

\paragraph{Solver.} Both the two- and three-dimensional
simulations solved the incompressible Navier--Stokes equations with the
open-source immersed-boundary solver \textsc{WaterLily}~\citep{waterlily}.
WaterLily is based on the Boundary Data
Immersion Method (BDIM): a body enters the simulation as a signed-distance
function together with its boundary velocity, which is how the solver couples
the solid into the flow. This makes it well suited to a continuously deforming,
self-propelled swimmer, whose moving boundary is simply re-measured at each
step as the body bends and translates. WaterLily is written in pure Julia and
runs on both CPU and GPU backends through KernelAbstractions.jl, making it a
fast and convenient base solver platform for our automated, many-evaluation
pipeline.

\paragraph{Body and spine.} We build the swimmer around a single bendable
backbone, illustrated in Extended Data Fig.~\ref{fig:extended-data-fluid-control}, and described by a normalised arc-length coordinate
\(s\in[0,1]\) running from the head (\(s=0\)) to the tail tip (\(s=1\)) and a
non-dimensional physical length \(L\). Two bending joints sit on this backbone, at \(s=1/3\) and
\(s=2/3\). Neither is a point hinge: each spreads its bend over a finite support
of half-width \(1/8\) in \(s\)---about a quarter of the body---so the spine
curves smoothly instead of kinking. Joint \(i\) contributes an incremental angle
\(\phi_i(t)\), and, read along the spine from the head, the local tangent angle
accumulates these contributions,
\begin{equation}
  \theta(s,t)=\sum_i \phi_i(t)\,H_i(s),
  \qquad
  \mathbf{C}(s,t)=L\!\int_0^s\!\big(\cos\theta,\ \sin\theta\big)\,\mathrm{d}s',
  \label{eq:broad-body-tangent}
\end{equation}
where each \(H_i(s)\) is a smooth (smootherstep) ramp climbing from \(0\) ahead
of its joint to \(1\) behind it, and integrating the tangent bends the whole
backbone into the spine centreline \(\mathbf{C}\). The two-dimensional body is
then fleshed out around the bent spine: at each station we step off the
centreline along its local normal \(\mathbf{n}(s,t)\) by a fixed half-thickness
\(w(s)\),
\begin{equation}
  \mathbf{X}_{\pm}(s,t)=\mathbf{C}(s,t)\pm L\,w(s)\,\mathbf{n}(s,t).
  \label{eq:broad-sweep}
\end{equation}
While the spine deforms as the controller presents, the head-to-tail thickness profile is fixed, thus the swimmer is formed by a prescribed thickness
distribution wrapped on a controllable, bendable spine.

For the three-dimensional simulations, we retained the same
two-joint centreline and essentially the same \(x\)--\(y\) top-view outline,
with only the snout half-width increased from \(0.02L\) to \(0.03L\) to avoid
an overly sharp nose. At each centreline station, the lateral profile was
extended into an elliptical section normal to the spine, with lateral
semi-axis \(Lw(s)\) and dorsoventral semi-axis \(L\rho(s)w(s)\). The
height-to-width ratio \(\rho(s)\) increased smoothly from \(0.75\) at the
flattened head to \(1.05\) over the trunk, with a transition scale of
\(0.18L\). The baseline height scale was one, and no separate caudal fin was
  included.

\paragraph{Actuation.} The body is deliberately underactuated: a
controller may act only through these two joints, and only by commanding their
angular accelerations,
\begin{equation}
  \mathbf{a}_\phi=(\ddot\phi_1,\ \ddot\phi_2),
  \label{eq:broad-action}
\end{equation}
which the testbed clips and integrates to joint rates and angles under fixed
limits: an incremental bend of \(45^\circ\), a rate of
\(260^\circ\,\mathrm{s}^{-1}\), and an acceleration of
\(1800^\circ\,\mathrm{s}^{-2}\).
In practice, only the angular rate and angular acceleration reach the limits with regularity, reflecting the finite muscle-force capacity to overcome fluid--structure coupling and body inertia, respectively.

\paragraph{Free-swimming response.} The joints deform the body,
but neither its global trajectory nor its heading is prescribed. The complete
explicit partitioned coupling is summarized by the fluid--load--body-state
update cycle in Extended Data Fig.~\ref{fig:extended-data-fluid-control}b.
At the beginning of a step, the current flow and body supply the hydrodynamic
loads used by both the controller observation and the free-body update. After
the commanded joint accelerations have been integrated, the updated joint
configuration and free-body pose define the signed-distance field
\(d(\mathbf{x},t)\) and boundary velocity \(\mathbf{u}_b(\mathbf{x},t)\).
The solver then re-measures this moving boundary and advances the
incompressible velocity and pressure fields with BDIM; the new flow state
supplies the loads for the following cycle. At each step, pressure and viscous tractions are
integrated on the swimmer surface to obtain the hydrodynamic force
\(\mathbf{F}_h\) and the out-of-plane moment \(M_z\) about its current centre,
\begin{equation}
  \mathbf{F}_h=\int_{\partial\mathcal{B}}\mathbf{t}_h\,\mathrm{d}s,
  \qquad
  M_z=\int_{\partial\mathcal{B}}
  \left[(\mathbf{x}-\mathbf{r}_c)\times\mathbf{t}_h\right]_z\,\mathrm{d}s,
  \label{eq:broad-hydrodynamic-loads}
\end{equation}
where \(\mathbf{t}_h\) denotes the traction exerted by the fluid on the body,
including both pressure and viscous contributions. The solver supplies the
moving-boundary flow solution and these surface-load contributions; the
subsequent free-body integration is performed by our coupling routine.

The free-body states are the global centre position \(\mathbf{r}_c\), heading
\(\psi\), translational velocity \(\mathbf{U}_c\) and yaw rate \(\Omega\). To
prevent internal bending from spuriously translating this reference point, the
instantaneous centroid \(\mathbf{c}_d(t)\) of the deformed swept profile is
removed in the body-to-world mapping. Equivalently, a world point is queried in
the deforming body frame as
\begin{equation}
  \mathbf{q}=\mathbf{R}(-\psi)\left(\mathbf{x}-\mathbf{r}_c\right)
  +\mathbf{c}_d(t),
  \label{eq:broad-free-swim-map}
\end{equation}
where \(\mathbf{R}\) is the planar rotation matrix. Thus
\(\mathbf{r}_c\) tracks global free-body motion, whereas changes in
\(\mathbf{c}_d\) describe deformation within the body frame.

Because this is an explicit fluid--body coupling, we apply an added-mass
correction in the instantaneous body frame. The previous-step acceleration is
retained in the added-mass term to stabilize the load-to-motion feedback when
displaced-fluid inertia is comparable to body inertia. The correction is
applied to the swimmer's axial and lateral translation and yaw. Let
\(\mathbf{F}_b^n=\mathbf{R}(-\psi^n)\mathbf{F}_h^n\) and
\(\mathbf{a}_b^n=\mathbf{R}(-\psi^n)\mathbf{a}_c^n\). The axial and lateral
accelerations and the yaw acceleration are updated as
\begin{equation}
  a_{b,j}^{n+1}=
  \frac{F_{b,j}^{n}+m_{a,j}a_{b,j}^{n}}{m+m_{a,j}},
  \quad j\in\{\parallel,\perp\},
  \qquad
  \alpha^{n+1}=\frac{M_z^{n}+I_a\alpha^{n}}{I_z+I_a},
  \label{eq:broad-added-mass-update}
\end{equation}
where \(m\) and \(I_z\) are obtained by integrating the swept profile at
neutral buoyancy (\(\rho_b=\rho_f\)). For the partitioned correction, the
profile is represented by an equal-area ellipse with semi-major axis
\(a=L/2\) and semi-minor axis \(b=A/(\pi a)\), giving
\begin{equation}
  m_{a,\parallel}=\rho_f\pi b^2,
  \qquad
  m_{a,\perp}=\rho_f\pi a^2,
  \qquad
  I_a=\frac{\rho_f\pi}{8}\left(a^2-b^2\right)^2.
  \label{eq:broad-added-mass-properties}
\end{equation}
No additional linear or angular damping is introduced in this update. The new
accelerations are rotated back to the world frame; velocity and yaw rate are
advanced explicitly, while centre position and heading use the trapezoidal
updates
\begin{equation}
  \begin{aligned}
    \mathbf{U}_c^{n+1}&=\mathbf{U}_c^n+\mathbf{a}_c^{n+1}\Delta t,
    &\quad
    \mathbf{r}_c^{n+1}&=\mathbf{r}_c^n+
    \frac{\mathbf{U}_c^n+\mathbf{U}_c^{n+1}}{2}\Delta t,\\
    \Omega^{n+1}&=\Omega^n+\alpha^{n+1}\Delta t,
    &
    \psi^{n+1}&=\psi^n+\frac{\Omega^n+\Omega^{n+1}}{2}\Delta t.
  \end{aligned}
  \label{eq:broad-pose-update}
\end{equation}
The updated pose and joint state rebuild the boundary for the next solver
step, closing the loop shown in Extended Data Fig.~\ref{fig:extended-data-fluid-control}b. Forward propulsion and
turning are therefore emergent consequences of the coupled pressure and
  viscous loads, rather than commanded quantities.

\paragraph{Two-dimensional task and state feedback.}\mbox{}\par
The two-dimensional episodes run in a four-cylinder wake within a
\(24L\times16L\) domain (\(1536\times1024\) cells at \(L=64\)) at a reference
Reynolds number \(Re=UL/\nu=1000\).
Here \(U=1\) is the reference velocity and \(L\) the swimmer
length; the kinematic viscosity was set to \(\nu=UL/Re\) in both two and
three dimensions.
The swimmer is released from a shared prewarmed wake state near the
right of the domain and must bring its head within a capture radius of
\(0.75L\) of a target.
The evolutionary scores reported for the two- and
three-dimensional tasks (Figs.~\ref{fig:structured-evolution}a and
\ref{fig:three-dimensional-adaptation}c) used the same untransformed
distance--time integral,
\begin{equation}
  s=-\int_0^1[d_h(\tau T)/L] \, \mathrm{d}\tau,
  \label{eq:navigation-score}
\end{equation}
where \(d_h\) is the planar head-to-target distance, \(T\) is
the episode horizon and \(\tau=t/T\). After early capture, the terminal
distance was held for the remaining horizon; higher scores therefore indicate
trajectories that remain closer to the target.
To facilitate optimization in the two-dimensional task, agent evolution and DRL
training used the same augmented objective,
\begin{equation}
  R = s-0.2\frac{d_h(t_{\mathrm{end}})}{L}
  +0.2\min\!\left(\frac{t_{\mathrm{end}}}{T},1\right)
  +2\mathbf{1}_{\mathrm{reach}}-2\mathbf{1}_{\mathrm{unstable}},
  \qquad T=300,
  \label{eq:optimization-score}
\end{equation}
where $t_{\mathrm{end}}$ is the episode duration and the indicators denote
target capture and numerical instability. All navigation-score plots report
the unaugmented distance--time integral $s$ to provide a consistent measure of
task performance.
Once the retained policies were obtained, we further evaluated their generalization
across a matrix of parameter variations without retraining.
We therefore define a rich input space rather than a single error term,
including an eight-sample history of body-frame target geometry and
trends, hydrodynamic force and yaw moment, body-frame velocity, joint state,
local flow and cylinder geometry.
During evolution, the policy is free to draw on any subset of these inputs to
construct its control logic.

\paragraph{Problem without solution class.} The testbed fixes only the body geometry, the two actuated joints with their angle/rate/acceleration limits, the state input, and the task itself; it prescribes no gait, no waveform, and no parameter space. This sets the problem apart from the common setting in which a swimmer
rides a fixed travelling-wave template and a controller chooses only among a
small set of wave parameters~\citep{maertens2017, eloy2013}, such as a discrete tail-beat amplitude or
prescribed propulsion cycle. Starting from a minimal seed policy, the entire control law \(\pi\), mapping observed state to joint acceleration \(\mathbf{a}_\phi\), is handed to an external solver as arbitrary interpretable source code, with no template given or fixed. The actuation interface is deliberately limited to two joint accelerations, so the design freedom resides in the state-dependent control strategy. The narrow actuation channel concentrates the principal design challenge in the state-dependent strategy, which is precisely the object of the agent's search.

\paragraph{Three-dimensional task.} We used the WaterLily/BDIM solver in three
dimensions with \(L=64\) grid units and \(Re=1000\) in still water. A
\(4L\times3L\times1.5L\) computational window
followed the swimmer within a \(24L\times16L\) virtual world.
The swimmer state was integrated in world coordinates. When
its centre moved four grid cells from the window anchor, the window origin
shifted by an integer number of cells in \(x\) and \(y\) and remained fixed in
\(z\). The stored velocity and pressure fields were translated by the opposite
integer offset, and newly exposed cells were initialized as quiescent fluid.
This construction preserves the inertial governing equations and transports
the discrete flow field without interpolation. The fluid and body geometry
were resolved in three dimensions, whereas rigid-body motion was limited to surge,
sway and yaw in the horizontal mid-plane; heave, roll and pitch were locked.
The navigation layout was defined in the \(x\)--\(y\) top-view plane, matching
the two-dimensional case. Because the three-dimensional test was conducted in
still water, no flow prewarming was required. We ran three independent
experiments from the minimal seed and three from a
retained two-dimensional policy; the latter initialized continued evolution in
three dimensions and was not evaluated as a frozen zero-shot transfer. Each
experiment comprised 40 iterations with four solver-driven and four
optimizer-driven proposals per iteration. The
three-dimensional results were reported using the same navigation score $s$ in
Eq.~\eqref{eq:navigation-score}. A
complete 40-iteration experiment was the independent unit; its candidate
policies form correlated search history and were not treated as independent
replicates.

\subsection{Behavior-cloned PPO comparator}
\label{sec:methods-drl}

We used behavior-cloned proximal policy optimization (BC--PPO) as the deep
reinforcement learning comparator. Clipped PPO~\citep{schulman2017ppo} was
implemented in Stable-Baselines3~\citep{raffin2021stablebaselines3}. Training
used the same nominal two-dimensional CFD--FSI environment, initial
configuration and actuation limits as the agent experiments
(see \hyperref[sec:methods-dogfish-simulation]{simulation setup}). The observation encoded
navigation, body and joint states, hydrodynamic loads, local flow and cylinder
geometry, together with the target history defined in that section. Separate
actor and critic networks each contained two fully connected layers of 128
units with tanh activations. The actor parameterized a diagonal Gaussian over
the two joint-acceleration commands; actions were clipped and scaled to the
shared physical limits and held for four adaptive CFD steps.

\paragraph{Propulsive initialization.}
To align the propulsive starting point across methods, we initialized the actor by
behavior cloning the same seed controller. This controller depended only on
joint angles and rates, providing propulsion without target-directed steering.
For each training run, we collected 32,768 state--action pairs with Gaussian
perturbations applied to the executed actions; unperturbed seed outputs
provided the training labels. The actor mean was fitted by minimizing
mean-squared action error with Adam. Observation-normalization statistics were
estimated from the demonstration data and held fixed during PPO. After
initialization, the neural policy supplied the complete joint command during
both training and evaluation.

\paragraph{Training objective and optimization.}
Training rewards were constructed from the augmented objective $R$ in
Eq.~\eqref{eq:optimization-score}. Distance rewards were distributed along
each trajectory, with a terminal correction ensuring that their undiscounted
sum equalled $R$. PPO applied temporal discounting during training; reported
navigation performance used the unaugmented score $s$ in
Eq.~\eqref{eq:navigation-score}. Rewards were not normalized.

PPO collected 512 policy transitions from each of four synchronous
environments per update. Each update used ten optimization epochs with
minibatches of 64, an Adam learning rate of $3\times10^{-5}$, a clipping range
of 0.2 and a target Kullback--Leibler divergence of 0.03. The discount and
generalized-advantage factors were $\gamma=0.9999^4$ and
$\lambda=0.9995^4$, respectively, expressed per policy transition for the
four-step action repeat. The initial standard deviation of the normalized
actions was 0.05, and no entropy bonus was applied. Each of the five reported
training runs comprised 600 episodes. High-scoring BC--PPO policies that had
reached the target in the nominal environment were held fixed during
generalization testing, without retraining or case-specific tuning.\par

\subsection{AI use declaration}
The use of language-model agents for controller design is described above and
in Supplementary Section~B, which reports the model, execution settings and
instructions. Generative AI was also used to assist manuscript drafting and
revision. OpenAI Codex and Google NotebookLM assisted with text revision;
OpenAI Codex also assisted with modifications to figure-generation scripts
under author direction. The authors retain
responsibility for the scientific interpretation, verification of AI-assisted
outputs and the final manuscript.

\section*{Acknowledgements}
We acknowledge NUS IT's Research Computing group for providing computational support.

\section*{Funding}
This work was supported by the National Research Foundation, Singapore, under the NRF Fellowship (Project No. NRF-NRFF17-2025-0006) to L.Y.

\section*{Author contributions}
\textbf{B.S.:} Conceptualization; Data curation; Formal analysis; Investigation;
Methodology; Software; Validation; Visualization; Writing---original draft;
Writing---review \& editing.
\textbf{W.G.:} Data curation; Formal analysis; Methodology; Software;
Validation; Visualization; Writing---original draft; Writing---review \& editing.
\textbf{Z.Y.:} Conceptualization; Formal analysis; Methodology; Software;
Writing---original draft; Writing---review \& editing.
\textbf{L.Y.:} Conceptualization; Funding acquisition; Methodology;
Project administration; Resources; Supervision; Writing---review \& editing.

\section*{Competing interests}
The authors declare no competing interests.

\section*{Data availability}
Source data underlying the main and Supplementary figures and tables are
provided at
\url{https://github.com/scaling-group/agent-fluid} under the Apache License 2.0.

\section*{Code availability}
Code for controller generation with self-evolving agents, CFD--FSI simulation, the BC--PPO comparator,
frozen-policy evaluation and figure reproduction is publicly
available at
\url{https://github.com/scaling-group/agent-fluid} under the Apache License 2.0.

\begingroup
\setlength{\bibsep}{6pt}
\interlinepenalty=10000
\makeatletter
\renewcommand{\@biblabel}[1]{#1.}
\makeatother
\ifdefined\arxivversion

\else
\bibliography{references}
\fi
\endgroup

\ifdefined\arxivversion
\end{bibunit}
\fi

\newcounter{extendeddatafigure}
\clearpage
\begingroup
\makeatletter
\let\c@figure\c@extendeddatafigure
\makeatother
\renewcommand{\thefigure}{\arabic{extendeddatafigure}}
\renewcommand{\theHfigure}{extendeddata.\arabic{extendeddatafigure}}
\renewcommand{\figurename}{Extended Data Fig.}
\pdfbookmark[1]{Extended Data Fig. 1}{extended-data-figure-1}

\begin{figure}[H]
  \centering
  \includegraphics[width=\linewidth]{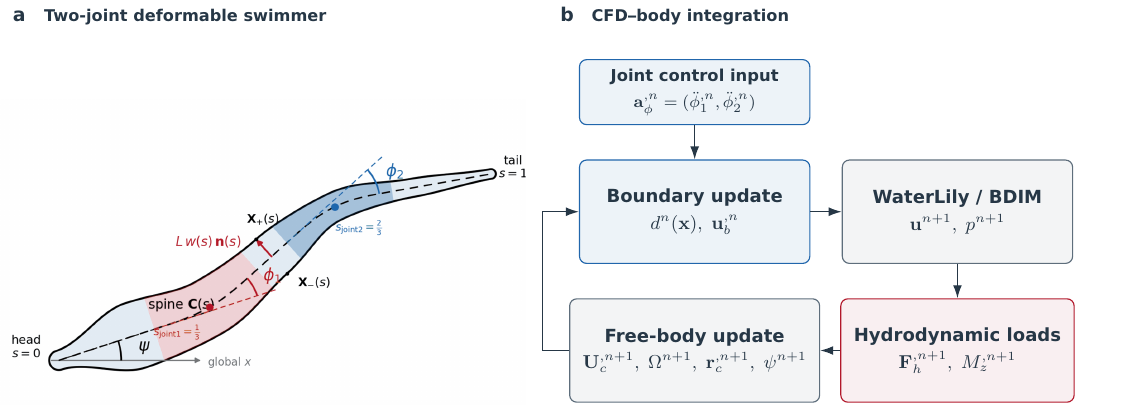}
  \caption{\textbf{Deformable swimmer and CFD--body
  integration.} (a)~A fixed thickness profile swept around a smooth two-joint
  spine. (b)~Joint angular-acceleration commands and the current pose update
  the moving boundary; WaterLily/BDIM advances the flow, surface loads drive
  free translation and yaw, and the updated pose closes the loop.}
  \label{fig:extended-data-fluid-control}
\end{figure}

\endgroup

\ifdefined\arxivversion
\clearpage
\setcounter{secnumdepth}{3}
\begin{bibunit}[sn-nature]
\begin{center}
  {\Large\bfseries Supplementary Information\par}
  \vspace{0.7em}
  {\large\bfseries Self-Evolving Scientific Agent Designs\\
  Physically Reasoned White-Box Fluid Control\par}
  \vspace{0.7em}
  Boai Sun, Wenjin Guo, Zongmin Yu and Liu Yang\par
  \vspace{0.3em}
  National University of Singapore
\end{center}
\phantomsection
\addcontentsline{toc}{section}{Supplementary Information}

\setcounter{section}{0}
\renewcommand{\thesection}{\Alph{section}}
\renewcommand{\thesubsection}{\thesection.\arabic{subsection}}
\renewcommand{\thesubsubsection}{\thesubsection.\arabic{subsubsection}}
\renewcommand{\theHsection}{supp.\Alph{section}}
\renewcommand{\theHsubsection}{\theHsection.\arabic{subsection}}
\renewcommand{\theHsubsubsection}{\theHsubsection.\arabic{subsubsection}}

\setcounter{figure}{0}
\renewcommand{\thefigure}{\arabic{figure}}
\renewcommand{\theHfigure}{supp.\arabic{figure}}
\renewcommand{\figurename}{Supplementary Figure}

\setcounter{table}{0}
\renewcommand{\thetable}{\arabic{table}}
\renewcommand{\theHtable}{supp.\arabic{table}}
\renewcommand{\tablename}{Supplementary Table}

\setcounter{equation}{0}
\renewcommand{\theequation}{S\arabic{equation}}
\renewcommand{\theHequation}{supp.\arabic{equation}}

\newtcolorbox{agentinstructionbox}[1]{
  breakable,
  sharp corners,
  colback=white,
  colframe=black!55,
  colbacktitle=black!5,
  coltitle=black,
  boxrule=0.45pt,
  left=7pt,
  right=7pt,
  top=5pt,
  bottom=6pt,
  before skip=9pt,
  after skip=11pt,
  fonttitle=\small\bfseries,
  fontupper=\small,
  toptitle=3pt,
  bottomtitle=3pt,
  title={#1},
  title after break={#1\space\textnormal{(continued)}}
}

\section{CFD verification and validation}
\label{sec:supp-cfd-validation}

We evaluated the numerical components used by this study independently of the
learned controller. The evidence is organized in three layers: an external
two-dimensional moving-boundary validation, an external three-dimensional
moving-boundary and power-integration validation, and an implementation check
of the three-dimensional moving-window algorithm.

\subsection{Two-dimensional wake-transition validation}
\label{sec:supp-2d-wake-validation}

We validated the two-dimensional moving-boundary solver
against the pure-pitch, leading-edge-pivot NACA~0016 benchmark of Lagopoulos,
Weymouth and Ganapathisubramani~\citep{lagopoulos2019transition} at
\(Re_C=1173\). Across seven Strouhal numbers, the predicted
BvK-to-reversed-BvK transition at \(N_c=C/\Delta x=64\) lay within 10\% of the
reported wake-neutral boundary (\(U_{\mathrm{wake}}\approx U_\infty\);
Supplementary Fig.~\ref{fig:supp-2d-validation}a).

For the same foil at \(Sr=0.25\) and \(A_D=0.874\), we then
refined the resolution from \(N_c=16\) to 128. The cycle-mean thrust
coefficient and propulsive efficiency, evaluated following Wang et
al.~\citep{wang2024learntoflap}, both converged with increasing resolution
(Supplementary Fig.~\ref{fig:supp-2d-validation}b). Balancing numerical
accuracy and computational cost, we therefore selected \(N_c=64\) for the
two-dimensional calculations.

\begin{figure}[H]
  \centering
  \includegraphics[width=\linewidth]{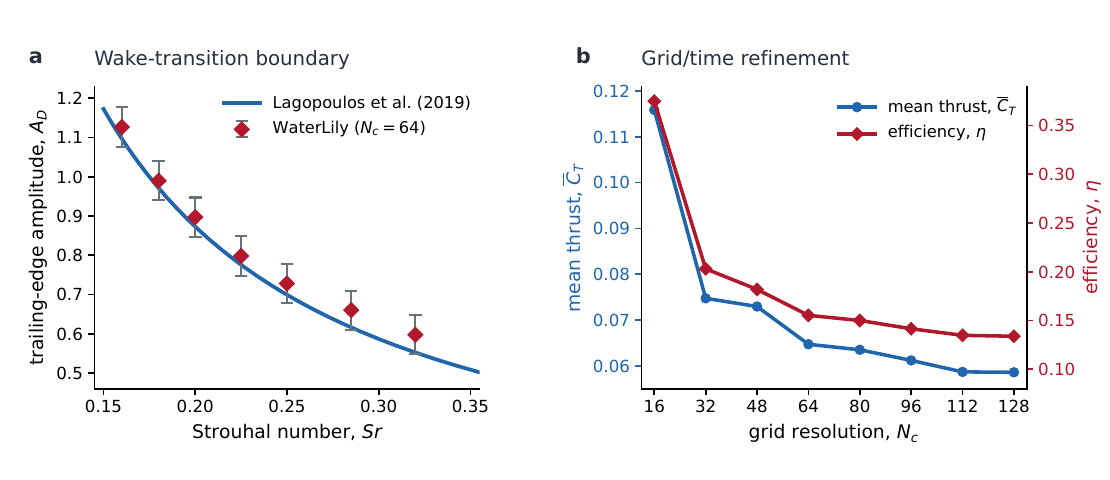}
  \caption{\textbf{Two-dimensional CFD validation.}
  \textbf{a}, Predicted BvK-to-reversed-BvK transition at \(N_c=64\) compared
  with the wake-neutral boundary reported by Lagopoulos et
  al.~\citep{lagopoulos2019transition}; bars show \(A_D\pm0.05\).
  \textbf{b}, Convergence of mean thrust and propulsive efficiency with
  increasing resolution.}
  \label{fig:supp-2d-validation}
\end{figure}
\FloatBarrier

\subsection{Three-dimensional oscillating-cylinder validation}
\label{sec:supp-3d-cylinder-validation}

We validated the three-dimensional solver against the
oscillating-cylinder benchmark published with
WaterLily~\citep{waterlily,waterlilyOscCylDriver} at \(Re_D=7620\),
\(V_r=5.4\), \(A_x/D=0.4\), \(A_y/D=1.6\) and \(\theta=\pi/6\). The domain
contained \((4n,2n,n)\) cells, with \(D/\Delta x=n/6\). Using the published
sign convention \(C_P=-(C_xu/U+C_yv/U)\), we compared the mean signed power
across nine resolutions with the reported numerical value \(-4.39\) and
experimental value \(-4.52\)~\citep{aktosun2024wakebranches}.

\begin{figure}[htbp]
  \centering
  \includegraphics[width=0.50\linewidth]{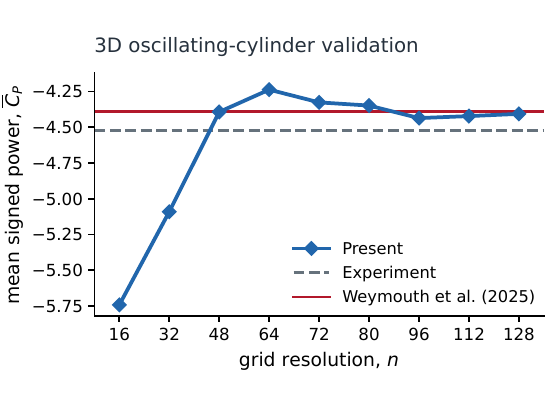}
  \caption{\textbf{Three-dimensional oscillating-cylinder validation.}
  Mean signed power coefficient across nine grid resolutions compared with
  the numerical value reported by Weymouth et al.~\citep{waterlily} and the experimental value.}
  \label{fig:supp-osc-cylinder-convergence}
\end{figure}
\FloatBarrier

With increasing resolution, the mean signed power approaches the published and
experimental values and remains close across the finer grids, demonstrating
numerical convergence. At \(n=64\), the deviations from both references are
already below 10\%. Balancing this accuracy against the cost of repeated
three-dimensional simulations, and using the same nominal resolution parameter
of 64 as in the two-dimensional study, we therefore selected \(n=64\) for the
production calculations.

\subsection{Three-dimensional moving-window implementation verification}
\label{sec:supp-moving-window-verification}

At the production resolution \(L=64\), a fixed computational domain of
\(24L\times16L\times1.5L\) contains approximately \(1.51\times10^8\) fluid
cells. To make
repeated three-dimensional closed-loop swimming simulations computationally
tractable, we use a compact \(4L\times3L\times1.5L\) moving window that follows
the swimmer while retaining the same local resolution and
fluid--structure/control configuration. The moving window contains
approximately \(4.72\times10^6\) cells, corresponding to 3.13\% of the fixed
full field and a 96.9\% reduction in active cell count.

We quantified the effect of this reduction using five matched pairs of
three-dimensional swimming simulations at \(Re=1000\). The fish was released
from five distinct locations and headings around a common target, including
approaches from each quadrant and from the western centreline. Within each
pair, the body geometry, initial joint state, frozen feedback policy, target,
time-step restriction and initially quiescent fluid were identical; only the
domain treatment differed. Every comparison was continued
for \(10T\), and the moving-window cases underwent 69--86
integer-cell translations.

At \(t=10T\), the southwest pair provides a same-time visual
comparison of the local wake, swimmer position and accumulated trajectory
after 71 window translations (Supplementary
Fig.~\ref{fig:supp-moving-window-comparison}). The trajectory, heading and
velocity differences across all five matched pairs, together with the relative
computational cost, are summarized in Supplementary
Table~\ref{tab:supp-moving-window-verification}.

\begin{figure}[H]
  \centering
  \includegraphics[width=\linewidth]{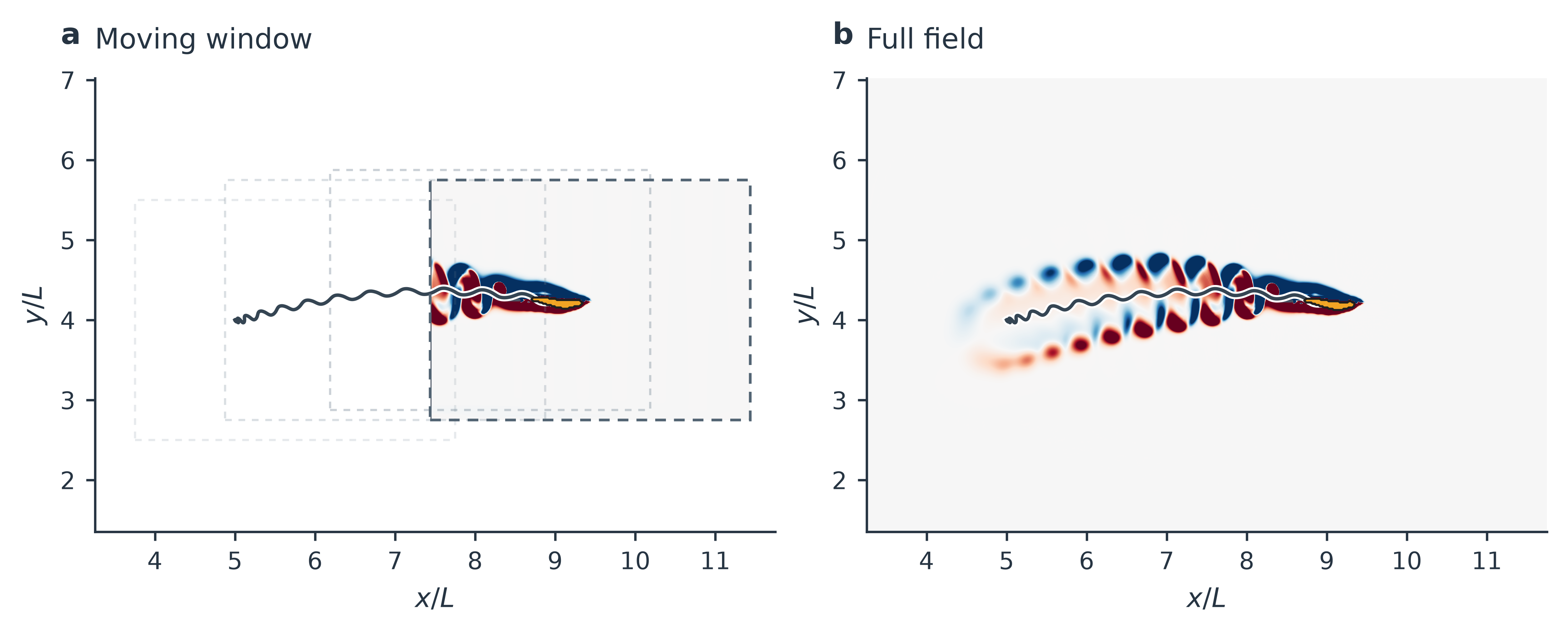}
  \caption{\textbf{Matched moving-window and full-field
  simulation.} Mid-plane vorticity and swimmer-centre trajectories at
  \(t=10T\) for the same southwest release, rendered directly from raw solver
  output with common axes, vorticity limits and local resolution. \textbf{a},
  Moving-window solution. Faded outlines trace preceding window positions, and
  the dark dashed box marks the current \(4L\times3L\) window. \textbf{b}, Fixed
  full-field solution over the same range. Both simulations used the same
  initial condition and frozen policy.}
  \label{fig:supp-moving-window-comparison}
\end{figure}

\begin{table}[htbp]
  \centering
  \small
  \setlength{\tabcolsep}{5pt}
  \renewcommand{\arraystretch}{1.12}
  \begin{tabular}{@{}lrrrrr@{}}
    \hline
    Release & Translations &
    \(\mathrm{RMS}(\Delta r_c)/L\) &
    \(\mathrm{RMS}(\Delta\psi)\) (deg) &
    \(\mathrm{RMS}(\Delta U_c)/U\) &
    Time ratio \\
    \hline
    Southwest & 71 & 0.00906 & 0.248 & 0.00662 & 19.6 \\
    Southeast & 84 & 0.0126 & 0.359 & 0.00576 & 20.1 \\
    Northwest & 86 & 0.0145 & 0.645 & 0.0107 & 22.0 \\
    Northeast & 69 & 0.0398 & 1.15 & 0.0166 & 19.8 \\
    West      & 76 & 0.0157 & 0.446 & 0.0104 & 22.4 \\
    \textbf{Mean} & \textbf{77.2} & \textbf{0.0183} &
    \textbf{0.570} & \textbf{0.0100} & \textbf{20.8} \\
    \hline
  \end{tabular}
  \caption{\textbf{Five-release matched full-field versus moving-window
  validation.} RMS differences are evaluated over \(0\leq t\leq10T\); the
  time ratio is the measured fixed-full-field/moving-window wall time per
  solver step.}
  \label{tab:supp-moving-window-verification}
\end{table}

\FloatBarrier

Across the five release configurations, the mean RMS differences in centre
position and heading were \(0.0183L\) and
\(9.95\times10^{-3}\,\mathrm{rad}\) (\(0.570^\circ\)), respectively. The mean
velocity RMS difference was \(0.0100U\). The maximum observed centre-position
and heading differences were \(0.0985L\) and
\(5.06\times10^{-2}\,\mathrm{rad}\)
(\(2.90^\circ\)). The fixed full field required 20.8 times the wall time per
solver step on average.

Because the policy is evaluated in closed loop from the instantaneous state
\(s(t)\), the controller continuously adapts its actions to the realised
fluid--swimmer state; the modest accumulated trajectory separation is thus
fully compatible with robust state feedback. The five-release results show
that the moving window preserves the relevant controlled swimming dynamics
while using 32-fold fewer active cells and approximately 21-fold less wall time
per step. The domain reduction therefore does not materially affect the
exploration or assessment of control capability.

\section{Self-evolving agents: implementation, prompts and instructions}
\label{sec:supp-agent-config}

The experiments used our Evolving Ensemble of Agents (EvE)
framework with the Codex agent provider,
\texttt{gpt-5.6-sol}, \texttt{xhigh} reasoning effort and a maximum of
18 turns per agent rollout, with live web search enabled. Four agent
workspaces ran in parallel at each evolutionary iteration. Parent agents and
reference entries were sampled using rank-based softmax weights with
temperature 1.0; the parent policy was selected uniformly from the sampled
policy examples. Agent ratings were updated using scalar Elo with $K=32$
and an initial rating of 1500. The resolved experiment configurations and
software revisions are archived with the accompanying code.

Each agent received a launch prompt and was instructed to read the workspace
instructions and current rollout evidence before editing the controller. The
first box reproduces the launch prompt used in the reported runs. The following
boxes reproduce the relevant workspace and Knowledge Shelf instructions. We
removed only repeated text and implementation-specific completion checks.
Bracketed notes identify and explain each omission; all other wording is
unchanged.

\begin{agentinstructionbox}{Launch prompt}
Read \texttt{README.md} and current rollout evidence first. Produce exactly one
multi-wake target-policy candidate in \texttt{solver/}. Keep candidate-specific
reasoning in \texttt{logs/optimize/wake\_policy\_notes.md}, and add or materially
revise at least one evidence-backed reusable lesson in
\texttt{guidance/control\_experience.md} from the assigned parent, sampled solver
results, and inherited logs. If no positive lesson survives the evidence,
record a concrete negative result and what later workers should avoid or test.
The candidate is incomplete until both files are updated. Work only inside this
Phase 2 workspace.

Use the fish-control bookshelf through this structured transfer protocol:

1. Before the first architecture proposal from the common naive seed, consult
\path{guidance/skills/fish-control-primitives/SKILL.md} and its references.

2. On later iterations, consult it again when the inherited guidance indicates
three consecutive completed iterations with neither a new controller mechanism
nor a semantic improvement such as a new success, a better termination class,
or a meaningfully different useful trajectory.

3. Consultation is required at those moments; adopting a primitive is
optional. Do not use the shelf to justify scalar-only gain tuning.

4. If a source informs the edit, record a concise block in
\texttt{logs/optimize/wake\_policy\_notes.md}:

\begin{quote}
\ttfamily\small
bookshelf\_consulted: true\par
source\_domain:\par
source\_mechanism:\par
transferable\_invariant:\par
nontransferable\_details:\par
policy\_translation:\par
falsification:
\end{quote}

Translate only the invariant into normalized body-frame observations and the
two-joint state-feedback contract. Prefer one mechanism or one small compatible
combination. Published gains, species-specific kinematics, exact vortex phases,
and task-specific routes are not transferable answers.
\end{agentinstructionbox}

\begin{agentinstructionbox}{Workspace instructions}
Produce one improved fixed-morphology policy that drives the fish diagonally
through the current four-cylinder asymmetric staggered wake field to the target.
Preserve the public \texttt{target\_policy\_params} and \texttt{target\_policy}
contract. The worker may improve feedback structure or parameter schema when
evidence supports it, but must not edit morphology, WaterLily, the episode,
geometry, or scoring code. Every active propulsion and steering parameter is
owned by \texttt{target\_policy\_params}; the fixed episode configuration cannot
replace it.

The configured EvE workers run independently and each produces exactly one
candidate. Do not create sibling candidates in this workspace.

Before changing the policy, write the candidate-specific visual diagnosis and
policy hypothesis to \texttt{logs/optimize/wake\_policy\_notes.md}. Keep transient
reasoning there; logs are provenance and do not create an optimizer child.

[Repeated guidance-update requirements are omitted because they appear in the
launch prompt above.]

The new candidate's CFD evaluation runs only after this worker exits and becomes
evidence for a later generation. A solver candidate is incomplete until both
notes and durable guidance change.

[Candidate-file safeguards and automated schema checks are omitted because they
are implementation-specific.]

Do not run formal CFD yourself; EvE evaluates the candidate after the worker
exits. Do not ask the human for clarification.
\end{agentinstructionbox}

\begin{agentinstructionbox}{Knowledge Shelf instructions}
This shelf collects mechanisms discovered in biological swimming, robotic fish,
reduced-order theory, CFD, and wake-control studies. The source problem does not
need to use this fish, geometry, or objective. The useful object is an
explanatory invariant that can be translated into this lane's observations,
two-joint actuation, and physical constraints.

Read current keyframes and metrics before using the shelf. Then read
\texttt{references/primitives.md}; read \texttt{references/sources.md} when a
source family informs an edit. Treat the references as hypotheses, not commands.
Published gains, dimensional frequencies, species-specific envelopes,
world-frame routes, and exact vortex phases are not transferable answers.

Keep the lane contract:

\begin{itemize}
\item Edit only \texttt{candidate\_target\_policy.jl}.
\item Use normalized body-frame observations and state feedback.
\item Do not add explicit time, step count, random numbers, file I/O, mutable
global state, fixed cylinder coordinates, case identity, or a memorized route.
\item Prefer one compact mechanism or one small compatible combination so the
next rollout can test what the transfer contributed.
\item A shelf citation never overrides contradictory rollout evidence.
\end{itemize}

\vspace{0.3\baselineskip}
\noindent
The worker entrypoint specifies how much consultation and bookkeeping is
required for the current experiment variant.
\end{agentinstructionbox}

\newpage
At each iteration, four agents received the same parent policy and common task
context, but different inherited guidance. Each also received four policy
examples and four guidance examples selected according to their ranks, together
with scores, logs and selected CFD evidence. Diagnoses remained with the
corresponding candidate, whereas general lessons were passed to later
iterations. Automated checks limited changes to the controller and verified the
required outputs. Each run comprised the Seed and 20 iterations, with four
candidate evaluations per iteration, giving 81 CFD evaluations.

\section{Design lineage of the evolved controller}
\label{sec:supp-control-lineage}

Supplementary Table~\ref{tab:supp-control-lineage} compares the control
structures used at seven milestones in the evolution of the champion controller.
The selected policies correspond to the Seed and Iterations 1, 6, 7, 12, 18 and 19.

\begin{table}[H]
  \centering
  \small
  \setlength{\tabcolsep}{4pt}
  \renewcommand{\arraystretch}{1.15}
  \begin{tabular}{@{}
    >{\raggedright\arraybackslash}p{0.23\linewidth}
    >{\raggedright\arraybackslash}p{0.17\linewidth}
    >{\raggedright\arraybackslash}p{0.51\linewidth}@{}}
    \hline
    Control structure & Policy stage & Implemented feedback and role \\
    \hline
    Travelling body wave
    & Seed
    & A state-feedback oscillator drives the anterior joint, while the
      posterior joint follows with a phase lag. The Seed observes only joint
      state and has no target or flow input. \\ \hline

    Body-frame bearing guidance
    & Iteration 1
    & Bearing is converted into a signed, bounded steering command and
      distributed across the two joint accelerations without prescribing a
      world-frame route. \\ \hline

    Course-responsive authority allocation
    & Iteration 6
    & Body-frame forward and lateral velocity modify the bearing request, while
      large bearing error reserves actuator authority for redirection. \\ \hline

    Half-cycle asymmetry
    & Iteration 7
    & Joint phase redistributes steering toward the half-cycle already moving
      in the requested direction while preserving the travelling-wave carrier.
      \\ \hline

    Response-scheduled redirect burst
    & Iteration 12
    & A transient burst augments the base asymmetry while large bearing error
      remains unmet. Recent bearing response, assisting yaw moment and high
      joint speed withdraw only this surplus authority. \\ \hline

    Crossflow response
    & Iteration 18
    & Target-aligned relative crossflow replaces bearing history as the
      response signal used to release surplus redirect authority. \\ \hline

    Force-confirmed wake assistance
    & Iteration 19
    & Lateral force must co-sign target-aligned crossflow before wake assistance
      releases the burst. This champion remained best through Iteration~20. \\
    \hline
  \end{tabular}
  \caption{\textbf{Control structures in selected policies across the
  evolution.} The table summarizes what feedback
  each milestone uses and how it affects the controller.}
  \label{tab:supp-control-lineage}
\end{table}
\FloatBarrier

\begingroup
\renewcommand{\refname}{Supplementary references}
\setlength{\bibsep}{6pt}
\interlinepenalty=10000
\makeatletter
\renewcommand{\@biblabel}[1]{#1.}
\makeatother

\endgroup
\end{bibunit}
\fi

\end{document}